%% file: paper.tex
\documentclass[runningheads]{llncs}
\usepackage[T1]{fontenc}
\usepackage{graphicx,verbatim}
\usepackage{booktabs}
\usepackage{adjustbox}
\usepackage{array}
\usepackage{hyperref}
\usepackage{graphicx}
\usepackage[utf8]{inputenc}
\usepackage{amsmath}
\usepackage{multirow}
\usepackage{cite}
\usepackage{float} 
\usepackage{enumitem}
\usepackage{diagbox}
\usepackage{xcolor}
\usepackage{amsfonts} 
\usepackage{mwe}
\usepackage{caption}

\usepackage{amsmath}
\usepackage{amssymb}
\usepackage{subcaption}
\usepackage{mathrsfs}

\newcolumntype{x}[1]{>{\centering\arraybackslash\hspace{0pt}}p{#1}}

\begin{document}
\title{Active Learning for Deep Learning-Based Hemodynamic Parameter Estimation}
\titlerunning{Active Learning for Hemodynamic Parameter Estimation}
\author{
Patryk Rygiel\inst{1} \and
Julian Suk\inst{1} \and
Kak Khee Yeung\inst{2,3,4} \and 
Christoph Brune\inst{1} \and
Jelmer M. Wolterink\inst{1}
}
\authorrunning{P. Rygiel et al.}
%
\institute{
Department of Applied Mathematics, Technical Medical Centre, University of Twente, Enschede, The Netherlands \\ \email{\{p.t.rygiel,j.m.suk,c.brune,j.m.wolterink\}@utwente.nl}
\and
Department of Surgery, Amsterdam University Medical Center, Location University of Amsterdam, Amsterdam, The Netherlands
\and
Department of Surgery, Amsterdam University Medical Center, Location Vrije Universiteit Amsterdam, Amsterdam, The Netherlands
\and
Amsterdam Cardiovascular Sciences, Atherosclerosis \& Aortic diseases, Amsterdam, The Netherlands
}

\maketitle
\begin{abstract}
Hemodynamic parameters such as pressure and wall shear stress play an important role in diagnosis, prognosis, and treatment planning in cardiovascular diseases. 
These parameters can be accurately computed using computational fluid dynamics (CFD), but CFD is computationally intensive.
Hence, deep learning methods have been adopted as a surrogate to rapidly estimate CFD outcomes.
A drawback of such data-driven models is the need for time-consuming reference CFD simulations for training.
In this work, we introduce an active learning framework to reduce the number of CFD simulations required for the training of surrogate models, lowering the barriers to their deployment in new applications.
We propose three distinct querying strategies to determine for which unlabeled samples CFD simulations should be obtained. 
These querying strategies are based on geometrical variance, ensemble uncertainty, and adherence to the physics governing fluid dynamics.
We benchmark these methods on velocity field estimation in synthetic coronary artery bifurcations and find that they allow for substantial reductions in annotation cost.
Notably, we find that our strategies reduce the number of samples required by up to 50\% and make the trained models more robust to difficult cases.
Our results show that active learning is a feasible strategy to increase the potential of deep learning-based CFD surrogates.


\end{abstract}

\section{Introduction}
Computational fluid dynamics (CFD) is commonly used to model blood flow  \textit{in-silico} for the assessment of cardiovascular diseases (CVDs)~\cite{morris2015cfd}.
CFD simulations can provide hemodynamic markers that have been found to correlate with the development and progression of various CVDs~\cite{bruyne2008ffr,mutlu2023cfd}.
The classical CFD pipeline involves medical imaging, extraction of a 3D model of the vasculature of interest, and applying a CFD solver with - ideally - patient-specific boundary conditions for blood flow modelling.
While accurate, CFD solvers are computationally expensive, requiring hours on high-performance clusters to converge for a single vascular model and a single set of boundary conditions.

To overcome the limitations of CFD, in recent years, deep learning-based CFD surrogate models have been proposed to estimate hemodynamics orders of magnitude faster than conventional CFD~\cite{taebi2022aicfdsurvey,arzani2022aicfdsurvey}. 
These surrogate models include geometric deep learning (GDL)~\cite{bronstein2021gdl} methods that can operate on point clouds or meshes using PointNet++~\cite{qi2017pointnet} or Transformer~\cite{vaswani2023} architectures. 
Applications include the estimation of surface (e.g., wall shear stress~\cite{suk2024mesh,suk2024lab}) and volumetric (e.g., velocity~\cite{Li2021aorta,suk2023velocity,suk2024physics,suk2024operator,suk2024lab}, pressure~\cite{Li2021aorta,suk2024operator,rygiel2023,nannini2025benchmark}) hemodynamic fields on 3D vascular models.

However, training of GDL models requires diverse datasets of 3D shapes and corresponding CFD solutions.
Generating reference labels in large data sets can be very time-consuming.
Hence, while these methods let us shift the bulk of the computation from inference to training, they come with computational challenges in label generation.
Moreover, the need for large data sets might hamper the deployment of such models to new applications.
This challenge is not unique to this problem but is ubiquitous in other machine learning tasks, including medical image computing tasks such as classification and segmentation~\cite{litjens2017survey}.
Previous works in machine learning have shown how active learning (AL) can limit annotation costs while maintaining model quality~\cite{kumar2020alsurvey, settles2009al}.
In AL, a model's training set grows incrementally by labeling and including only those samples deemed most informative for training that model. 
The challenge in AL is to design a suitable query strategy based on metrics that estimate how informative a sample is, \textit{without} access to its true label.


In this work, we introduce an AL framework for deep learning-based CFD surrogate models (Fig.~\ref{fig:workflow}).
We propose and evaluate three distinct query strategies for AL that operate either in the shape (input) or hemodynamic (output) domain of the model~\cite{zhang2019uncertainty,arthurs2021activecfd}.
First, a query that considers the most informative sample to be that whose shape most differs from the already included training shapes. Second, a strategy that considers the output variance as measured in Monte Carlo dropout, akin to model entropy in classification-based AL approaches.
Third, we include samples for which the model's output most violates the underlying physics as described through the Navier-Stokes equations. 
Through a series of experiments on velocity field estimation in synthetic models of coronary artery bifurcations, we show how the proposed query strategies enable a reduction of annotation costs while improving the estimator's performance, especially its robustness towards difficult and outlier cases.
We observe that the task-oriented query strategy based on physics adherence is the most suitable choice, allowing for an almost two-fold reduction in annotated samples compared to other approaches.

\section{Methods}
\input{figures/workflow-fig}


We consider three distinct data pools: labeled $\mathcal{L}$, unlabeled $\mathcal{U}$, and test $\mathcal{T}$. 
Pools $\mathcal{L}$ and $\mathcal{T}$ contain samples $(x_i, y_i)$, where $x_i$ is a 3D shape representation and $y_i$ is a hemodynamics field.
However, for samples in the unlabeled pool $\mathcal{U}$, the hemodynamics fields are initially unknown.
For a shape $x_i \in \mathcal{U}$, the label $y_i$ can be obtained by querying an oracle $\mathcal{O} \colon x_i \mapsto y_i$.
In our work, the oracle $\mathcal{O}$ is a CFD engine.
 

We aim to train a model $F(x_i, \theta) = y_i$ that achieves the best performance on $\mathcal{T}$ while requiring a minimal number of calls to the oracle $\mathcal{O}$ with unlabeled samples from $\mathcal{U}$.
To do so, we iteratively choose the most informative unlabeled query pool $\mathcal{Q} \subset \mathcal{U}$ and obtain labeled query pool $\mathcal{Q}^{+} = \{(q_i, \mathcal{O}(q_i)): q_i \in \mathcal{Q}\}$ by querying the oracle $\mathcal{O}$.
Subsequently, the labeled pool $\mathcal{L}$ is extended with $\mathcal{Q}^{+}$ and the model $F$ retrained (Fig.~\ref{fig:workflow}). 
We use pool-based AL, with $|\mathcal{Q}| > 1$, which is common when dealing with deep learning models since retraining costs are high and individual samples only minimally influence model performance. 

\subsection{Query Strategies}
The main question we address in this paper is how to design a suitable query strategy for selecting $\mathcal{Q}$, i.e., samples for which CFD labels should be computed. 
We propose and evaluate three different query strategies adapted to the tasks of geometry processing and hemodynamics estimation.

\paragraph{Geometry-variance query (GV)} is a training-free querying strategy that selects the most geometrically varying samples for query pool $\mathcal{Q}$.
It is based on the assumption that different geometries provide distinctive information, allowing efficient sparse spanning of the input space.
As a geometric similarity measure, we employ the Chamfer distance between point sets $\mathcal{P}_1$ and $\mathcal{P}_2$ representing two different shapes $x_1$, $x_2$:

\begin{align*}
    &d_\text{Chamfer} (\mathcal{P}_1, \mathcal{P}_2) := \frac{1}{\lvert \mathcal{P}_1 \rvert} \sum_{p_1 \in \mathcal{P}_1} \underset{p_2 \in \mathcal{P}_2}{\text{min}} \lVert p_1 - p_2 \rVert_2 + \frac{1}{\lvert \mathcal{P}_2 \rvert} \sum_{p_2 \in \mathcal{P}_2} \underset{p_1 \in \mathcal{P}_1}{\text{min}} \lVert p_2 - p_1 \rVert_2&
\end{align*}

We compute all pairwise distances $d_\text{Chamfer}$ between samples in $\mathcal{L} \cup \mathcal{U}$ and construct a distance matrix $\mathbf{M}$ where $\mathbf{M}_{i,j}$ represents Chamfer distance between shapes $x_i, x_j \in \mathcal{L} \cup \mathcal{U}$.
The transductive variance of shape $x_i$ is then represented via transductive descriptor $\mathbf{M_i}$, the $i$-th row of distance matrix $\mathbf{M}$.
To construct query pool $\mathcal{Q}$, the most distinct shapes to the ones already in $\mathcal{L}$ are sampled through \textit{farthest-point-sampling}~\cite{eldar1997fps} with Euclidean distance in $\mathbb{R}^{|\mathcal{L} \: \cup \: \mathcal{U}|}$ on transductive descriptor set $\{\mathbf{M}_i: x_i \in \mathcal{U}\}$.


\paragraph{Query-by-committee (QBC)} is an uncertainty-based querying paradigm that uses a model ensemble, or \textit{committee}~\cite{burbridge2007qbc}.
The query pool $\mathcal{Q}$ is constructed by choosing the most ambiguous samples, where ambiguity is defined by comparing individual predictions of the committee members.
Here, instead of explicitly training an ensemble of models, we use \textit{Monte Carlo dropout}~\cite{gal2016mcdropout,zhang2019uncertainty} to approximate training multiple models. 
Velocity field prediction is a point-wise regression task, and hence we employ a mean over point-wise variance across the committee members to serve as a ranking metric:

\begin{align*}
    &\overline{\text{Var}}(Y)(x_i) := \underset{p \in \mathcal{P}_i}{\text{mean}} \ \text{Var}(Y_p) \ &(\text{variance})&
\end{align*}

where $Y$ is a random variable representing committee members' predictions and $\mathcal{P}_i$ is a point cloud representing the shape $x_i$. 

\paragraph{Physics-adherence query (PA)} quantifies the \textit{quality} of the predicted output in an unlabeled sample, defined by the extent to which the predicted velocity field adheres to the Navier-Stokes equations governing the fluid dynamics.
The momentum and continuity terms of the Navier-Stokes equation are computed over the predicted velocity field as follows~\cite{suk2024physics}: 

\begin{align*}
    L_\text{continuity}(x_i) &:= \underset{p \in \mathcal{P}_i}{\text{mean}} \lvert (\nabla \cdot y_i)^p \rvert \ &(\text{continuity}) \\
    L_\text{momentum}(x_i) &:= \underset{p \in \mathcal{P}_i}{\text{mean}} \lVert\rho((y_i \cdot \nabla) y_i)^p - \mu(\Delta y_i)^p \rVert_2 \ &(\text{momentum})
\end{align*}

where $\mathcal{P}_i$ is a point cloud representing the shape $x_i$,
$\Delta$ the Laplacian operator, and $\rho$
and $\mu$
the density and dynamic viscosity, respectively. 
Note that in the momentum term, we omit the pressure drop and accept it as a constant error, because we do not model pressure in this study.
We use the joint Navier-Stokes term $L_\text{NS}=L_\text{continuity}+\lambda L_\text{momentum}$ with $\lambda=1 \times 10^{-4}$ as a ranking metric.

\subsection{Architecture}
The task of the model $F$ is to predict volumetric velocity fields for 3D point clouds representing the lumen of (synthetic) artery models. We define $F(\cdot, \theta)$ using a PointNet++ architecture~\cite{qi2017pointnet}, which is a good model for hemodynamics estimation with relatively fast convergence time~\cite{nannini2025benchmark}.
Input features are, for each point, relative position to the closest arterial wall, inlet, and outlet~\cite{suk2023velocity}. 

\subsection{Quantitative Evaluation}
For all experiments, we utilize a dataset of $2,000$ synthetic left main coronary bifurcations geometries, in which steady-state, fixed boundary CFD simulations were performed~\cite{suk2023velocity}, taking around $15$ min per sample using SimVascular~\cite{updegrove2017simvascular}. 
Out of $2,000$ samples, the test pool $\mathcal{T}$ contains $1,000$ samples, and the remaining $1,000$ samples are used in the AL experiments. 
We use approximation disparity (Approx. disp.) and cosine similarity (Cos. similarity) to assess the accuracy of predicted velocity fields $\hat{y}$ w.r.t. ground-truth fields $y$ in the test set $\mathcal{T}$:

\begin{align*}
    \text{Approx. disp.} &:= \sqrt{\sum_{p \in \mathcal{P}}{\lVert \hat{y}^p - y^p \rVert^2_2 \ / \ \sum_{p \in \mathcal{P}}{\lVert y^p \rVert^2_2}}}& \\
    \text{Cos. similarity} &:= \underset{p \in \mathcal{P}}{\text{mean}} \ \text{cos}\angle(\hat{y}^p, y^p)&
\end{align*}

where $\mathcal{P}$ is a point cloud representing the input shape $x$.
Moreover, to evaluate adherence to the Navier-Stokes equations, we compute $L_\text{continuity}$ (continuity) and $L_\text{momentum}$ (momentum).

\section{Experiments \& Results}
\input{figures/boxplot-fig}
\input{figures/results-table}

The PointNet++ model used in all experiments is implemented in PyTorch~\cite{paszke2019pytorch} and PyTorch Geometric~\cite{fey2019pyg}.
All models are trained for $20,000$ iterations with a batch size of $2$, $\texttt{Adam}$ optimizer with a learning rate of $3e-4$ and an exponential learning rate scheduler with decay parameter $\gamma=0.9989$.
Training one model takes around two hours on a single NVIDIA A40/48G GPU.
Models are optimized using a combination of L1 loss for the velocity field magnitude and cosine similarity loss for its direction.

We select two random samples from $\mathcal{U}$, label them, and train our initial model on them.
Then, for each strategy, six rounds of AL are performed, with the size of the query pool increasing to $(4, 8, 16, 32, 64, 128)$. 
In each round, a new model is trained from scratch, with a dataset that includes the samples that were added based on the query strategy. 
We conduct each experiment five times with different initial samples.
We compare our querying strategies to a random choice of query pool $\mathcal{Q}$ - denoted as \textbf{Random}. 
Moreover, to get an upper bound on model performance, we train a model on all $1,000$ samples and denote it as \textbf{Baseline (1000)} in the following sections.

\input{figures/cummulative-fig}

\subsection{Overall Performance}
Figure~\ref{fig:boxplot} shows the progression of model performance through rounds of our AL experiment.
As may be expected, we observe a clear relation between the number of training samples included and model performance, where both approximate disparity and cosine similarity improve with more training data.
These results show that for the first four rounds of training (4, 8, 16, 32 training samples), randomly sampling from $\mathcal{U}$ leads to the largest improvement for both metrics. 
However, results obtained using these small data sets are still substantially worse than the baseline model trained with all samples, \textbf{Baseline (1000)}. 

In the last two rounds (64 and 128 training samples), this pattern is inverted, and random sampling no longer leads to the largest improvement in model performance. 
Instead, we observe that our query strategies perform significantly better across the board in comparison to the \textbf{Random}.
Table~\ref{tab:results-table} lists quantitative results for these two rounds.  
Among them, the PA query strategy is the one that offers the highest performance gain with respect to previous rounds.
Notably, models trained with this query strategy require only $64$ training samples whereas other strategies require $128$ training samples to reach the same performance. 
Moreover, we see a clear relationship between approximation disparity and physics-based metrics, \textit{continuity} and \textit{momentum}.

\subsection{Sample-Specific Performance}
The results in Fig.~ \ref{fig:boxplot} and Table~\ref{tab:results-table} indicate that model behavior across the sample distribution depends on the query strategy. 
To further investigate this relation, we plot the empirical cumulative distribution of approximation disparity (Fig.~\ref{fig:cumcorr} (a)).
We assume that easier samples have lower errors, while difficult ones have higher errors.
We find that for the 20-40\% easiest samples, the \textbf{Random} strategy performs better.
However, our query strategies lead to notably improved performance for more difficult cases at the cost of slightly reduced performance for easier samples. 
In fact, this cost is minimal, as illustrated in an easy example in Fig.~\ref{fig:visual}(b). 
Conversely, models trained using the proposed query strategies offer much more reliable predictions in the difficult sample~Fig.\ref{fig:visual}(a). 



Of our proposed query methods, the PA query strategy offers the best performance. 
This is supported by the results in Fig.~\ref{fig:cumcorr}(b), which show the correlation between the query ranking metrics and the approximation disparity computed for the \textbf{Baseline (1000)} model. 
These results indicate that the PA query provides an accurate estimate of the model's true performance on unlabeled cases, as required for successful AL.

\input{figures/qualitative-fig}



\section{Discussion \& Conclusion}
We have presented an active learning approach for deep learning-based hemodynamics estimation, which can substantially reduce the required number of training samples compared to random sampling. 

Query strategies in AL contain an \textit{exploration-exploitation} trade-off~\cite{settles2009al}. 
While querying at random is purely explorative, the other proposed strategies exhibit notions of exploitation either through input (GV), output variance (QBC) or output quality (PA). 
We have seen that exploitative queries are better at selecting \textit{difficult} unlabeled samples. 
This, in turn, leads to improved robustness to similarly difficult cases in testing scenarios.
However, since the annotation budget per round is fixed, fewer \textit{easy} samples are labeled and included in the training set. 
This leads to slightly worse performance on easy samples than in random sampling.
This phenomenon has previously been observed as the trade-off between model robustness and its natural accuracy~\cite{tsipras2019robustnessoddsaccuracy}. 
Hence, we establish that training with exploitative queries leads to more robust surrogate models than a random selection of samples.

We have explored applying the query strategies individually over the full active learning experiments.
However, the exploitation-based queries acknowledge different aspects of the data variance and quality.
These queries may provide complementary information, and further investigation is needed to determine whether combining the queries in a single active learning experiment yields better results than using them individually.

In this work, we have limited ourselves to velocity modelling in synthetic anatomies with fixed boundary conditions for CFD simulations.
As demonstrated in previous studies on hemodynamic modelling, the deep learning models can be readily extended to incorporate boundary conditions and predict both velocity and pressure~\cite{suk2024operator}.
A future step would be to apply the proposed strategies to active learning with real anatomical datasets with varying boundary conditions.
This scenario might be more challenging due to geometrical and hemodynamic variability, but the potential gains in labeling time would also increase substantially, from a few minutes for synthetic cases~\cite{suk2023velocity,rygiel2023} up to several hours for real ones~\cite{suk2024operator} per simulation that can be omitted.
Moreover, it remains to be investigated whether the current results are model-independent and if our findings would also hold for other (geometric) deep learning surrogate models than the PointNet++ used here. 

In conclusion, we have demonstrated how active learning with proposed querying strategies can improve the data efficiency and reliability of building AI-based biomedical CFD surrogates.

\subsubsection{Acknowledgments:}
This project has received funding from the European Union's Horizon Europe research and innovation programme under grant agreement No 101080947 (VASCUL-AID).

\newpage

\bibliographystyle{splncs04}
\bibliography{biblio}

\end{document}

%% file: figures/workflow-fig.tex
\begin{figure}[!t]
    \centering
    \includegraphics[width=\textwidth]{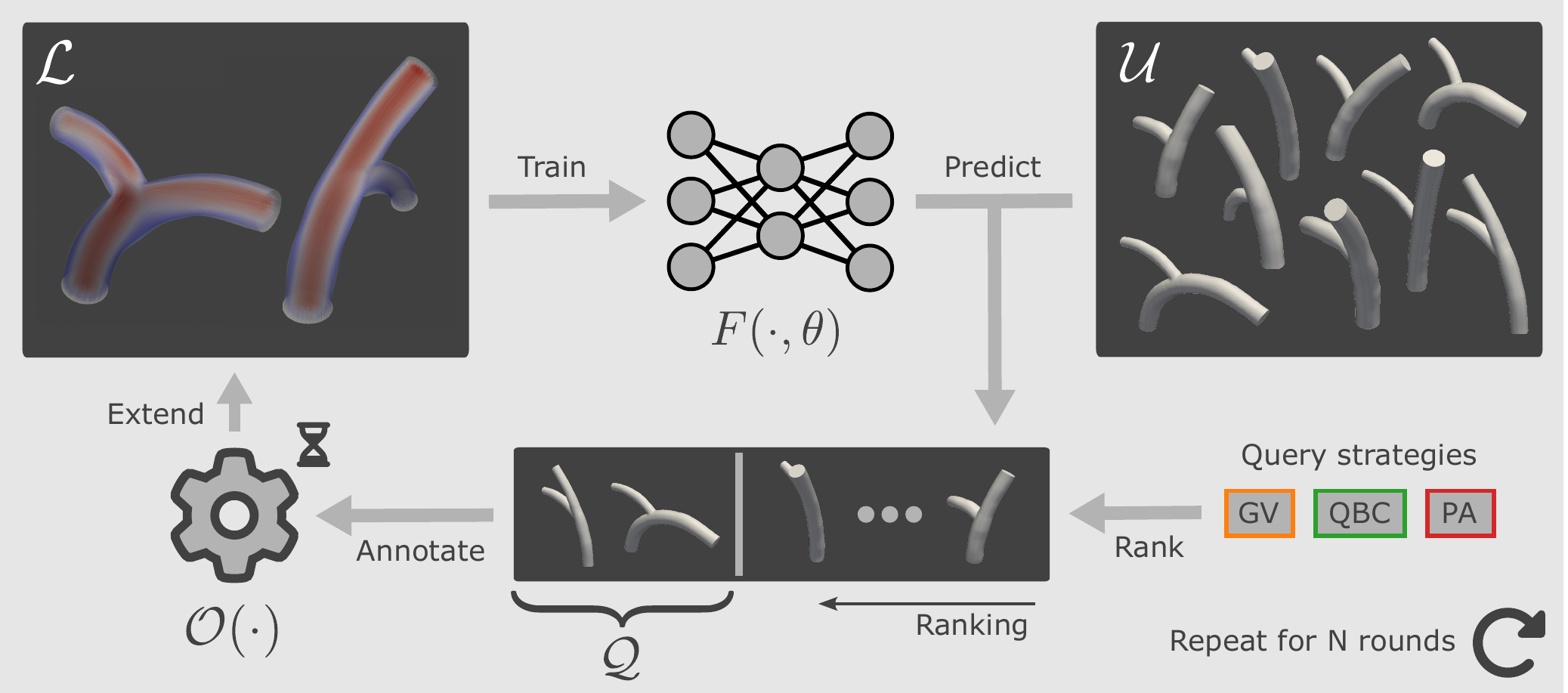}
    \caption{Active learning framework for deep learning-based CFD surrogate models. Model $F(\cdot, \theta)$ is trained on labeled pool $\mathcal{L}$, after which model predictions on unlabeled pool $\mathcal{U}$ are computed. Samples are ranked according to the query strategy, and the top $k$ samples comprise the query pool $\mathcal{Q}$. The oracle $\mathcal{O}(\cdot)$ - a CFD engine - labels all samples in $\mathcal{Q}$, which are then added to $\mathcal{L}$ for retraining of $F$. The procedure is repeated $N$ times or until a performance criterion is met.
    }
    \label{fig:workflow}
\end{figure}

%% file: figures/boxplot-fig.tex
\begin{figure}[!t]
    \centering
    \includegraphics[width=\textwidth]{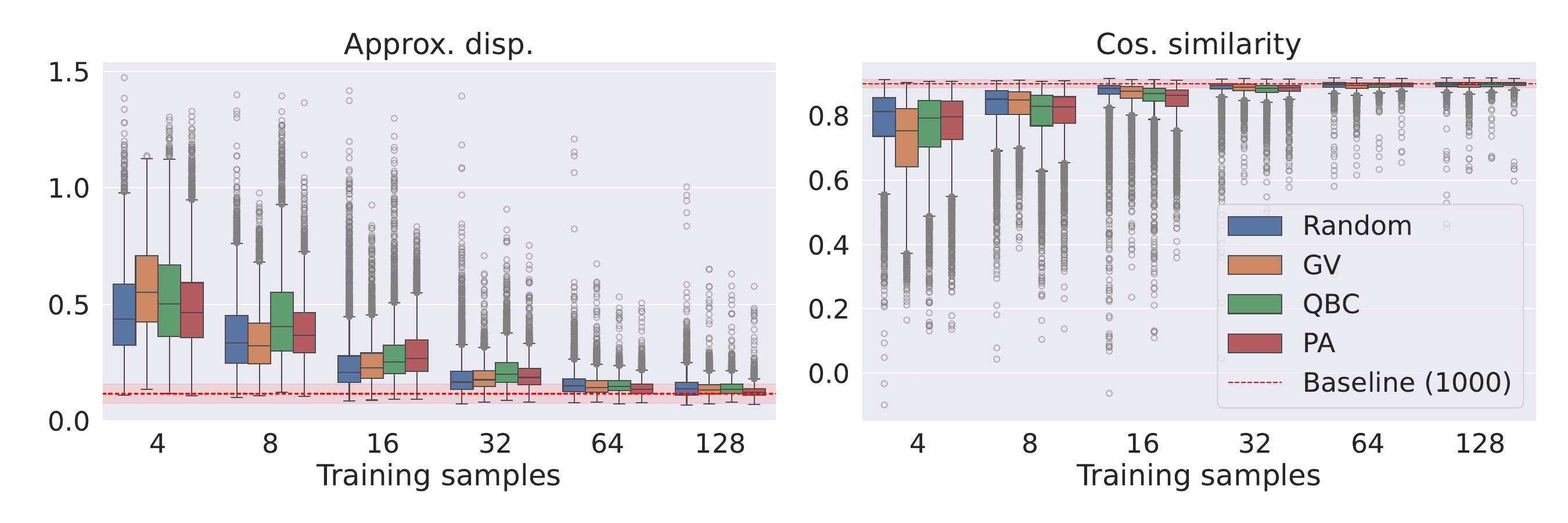}
    \caption{Approx. disp. and cos. similarity over the test pool $\mathcal{T}$ for different query strategies over all active learning rounds.}
    \label{fig:boxplot}
\end{figure}

%% file: figures/results-table.tex
\setlength{\tabcolsep}{0.5em}
{\renewcommand{\arraystretch}{1.2}%
    \begin{table}[!t]
    \centering
    \caption[Short Heading]{We report \texttt{mean} $\pm$ \texttt{std} in $\mathcal{T}$ for the two final rounds of the active learning experiments, with $64$ and $128$ training samples, respectively. Bold values indicate the best performance per metric per training set size.}
    \begin{adjustbox}{width=\textwidth}
    \begin{tabular}{llll|ll}
    \hline
    \toprule         
        \textbf{Query} & \textbf{\# training} & \textbf{Approx. disp} $\downarrow$ & \textbf{Cos. similarity} $\uparrow$ & \textbf{Continuity} 1e1 $\downarrow$ & \textbf{Momentum} 1e5 $\downarrow$\\
        

        \midrule
        \textbf{Random} & \multicolumn{1}{c}{\multirow{4}{*}{64}} & $0.161$ $\pm \ 0.061$ & $0.894$ $\pm \ 0.024$ & $0.952$ $\pm \ 0.285$ & $1.225$ $\pm \ 0.455$ \\ 
        \textbf{GV} & & $0.154$ $\pm \ 0.047$ & $0.891$ $\pm \ 0.017$ & $0.849$ $\pm \ 0.175$ & $0.956$ $\pm \ 0.140$\\ 
        \textbf{QBC} & & $0.155$ $\pm \ 0.036$ & $0.894$ $\pm \ 0.012$ & $0.819$ $\pm \ 0.130$ & $0.934$ $\pm \ 0.109$\\ 
        \textbf{PA} & & $\mathbf{0.143}^*$ $\pm \ 0.034$ & $\mathbf{0.896}$ $\pm \ 0.011$ & $\mathbf{0.763}^*$ $\pm \ 0.106$ & $\mathbf{0.896}^*$ $\pm \ 0.101$\\ 

        \midrule
        \textbf{Random} & \multicolumn{1}{c}{\multirow{4}{*}{128}} & $0.146$ $\pm \ 0.055$ & $0.897$ $\pm \ 0.018$ & $0.920$ $\pm \ 0.390$ & $1.214$ $\pm \ 0.444$\\ 
        \textbf{GV} & & $0.140$ $\pm \ 0.040$ & $0.894$ $\pm \ 0.016$ & $0.802$ $\pm \ 0.150$ & $0.907$ $\pm \ 0.113$\\ 
        \textbf{QBC} & & $0.141$ $\pm \ 0.035$ & $0.896$ $\pm \ 0.011$ & $0.796$ $\pm \ 0.124$ & $0.907$ $\pm \ 0.104$\\ 
        \textbf{PA} & & $\mathbf{0.128}^*$ $\pm \ 0.029$ & $\mathbf{0.898}^*$ $\pm \ 0.011$ & $\mathbf{0.745}^*$ $\pm \ 0.089$ & $\mathbf{0.886}^*$ $\pm \ 0.095$\\ 

        \midrule
        \textbf{Baseline} & \multicolumn{1}{c}{\multirow{1}{*}{1000}} & $0.117$ $\pm \ 0.042$  & $0.900$ $\pm \ 0.013$ & $0.729$ $\pm \ 0.130$ & $0.853$ $\pm \ 0.113$\\ 

        \bottomrule
        \multicolumn{6}{l}{\small * $p < 0.05$ in a Wilcoxon test between best and second best-performing query strategy.} \\
        
    \end{tabular}
    
    \end{adjustbox}
    \label{tab:results-table}
    \end{table}
}

%% file: figures/cummulative-fig.tex
\begin{figure}[!t]
    \centering
    \begin{subfigure}[t]{0.66\textwidth}
        \centering
        \includegraphics[width=\textwidth]{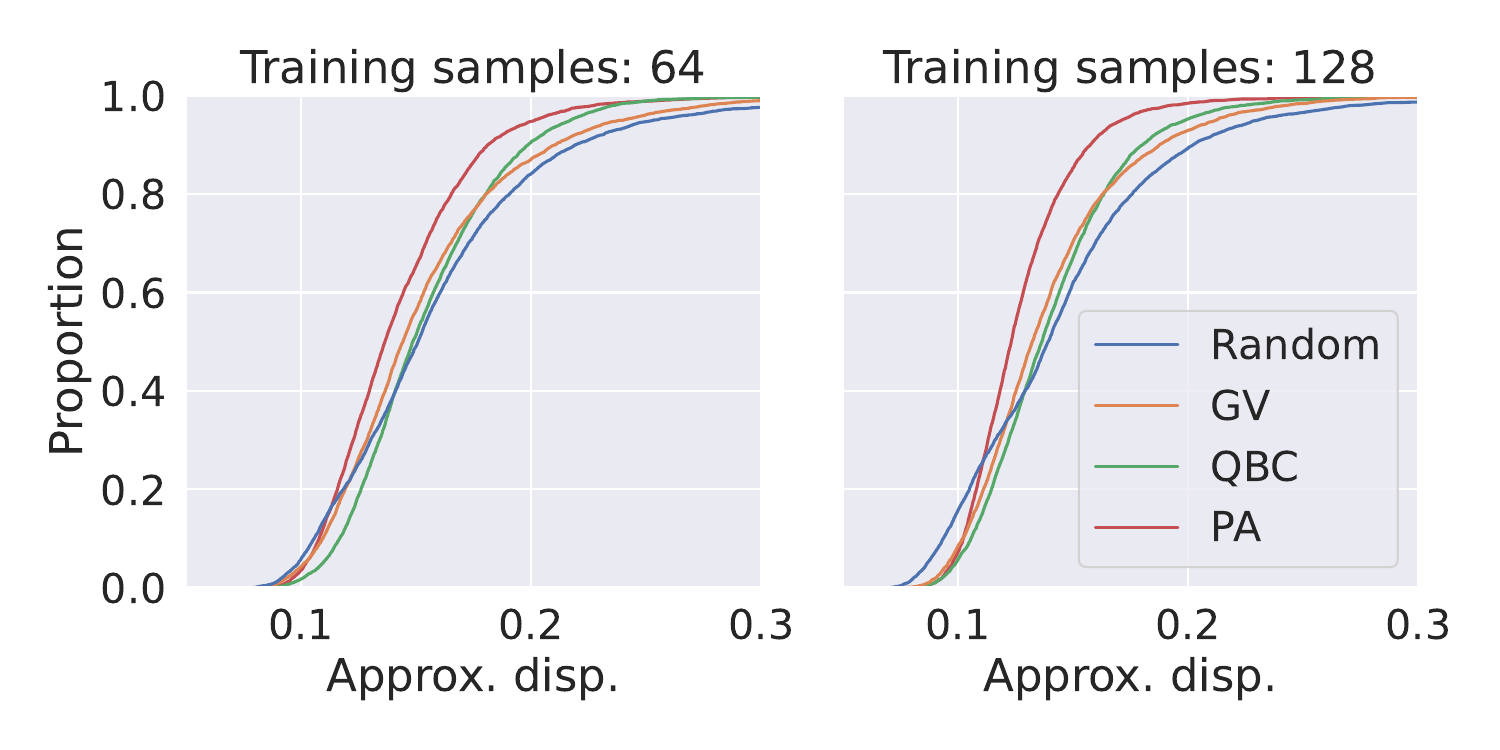}
        \caption{}
        \label{fig:cummulative}
    \end{subfigure}%
    ~
    \begin{subfigure}[t]{0.33\textwidth}
        \centering
        \includegraphics[width=\textwidth]{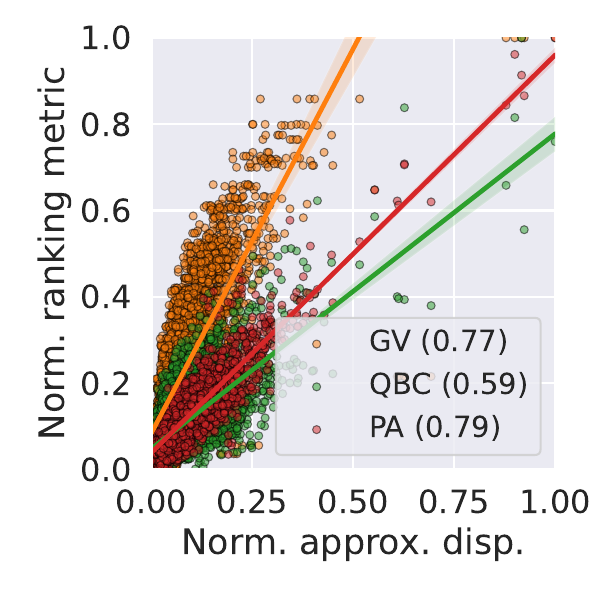}
        \caption{}
        \label{fig:correlation}
    \end{subfigure}
    \caption{Figure (a) showcases an empirical cumulative distribution plot of approximation disparity over the test pool $\mathcal{T}$ for the last $2$ rounds of active learning. Figure (b) showcases the correlation between normalized query ranking metrics and normalized approximation disparity. with Spearman correlation coefficient on test pool $\mathcal{T}$ for \textbf{Baseline (1000)}.}
    \label{fig:cumcorr}
\end{figure}

%% file: figures/qualitative-fig.tex
\begin{figure}[!t]
    \centering
    \includegraphics[width=\textwidth]{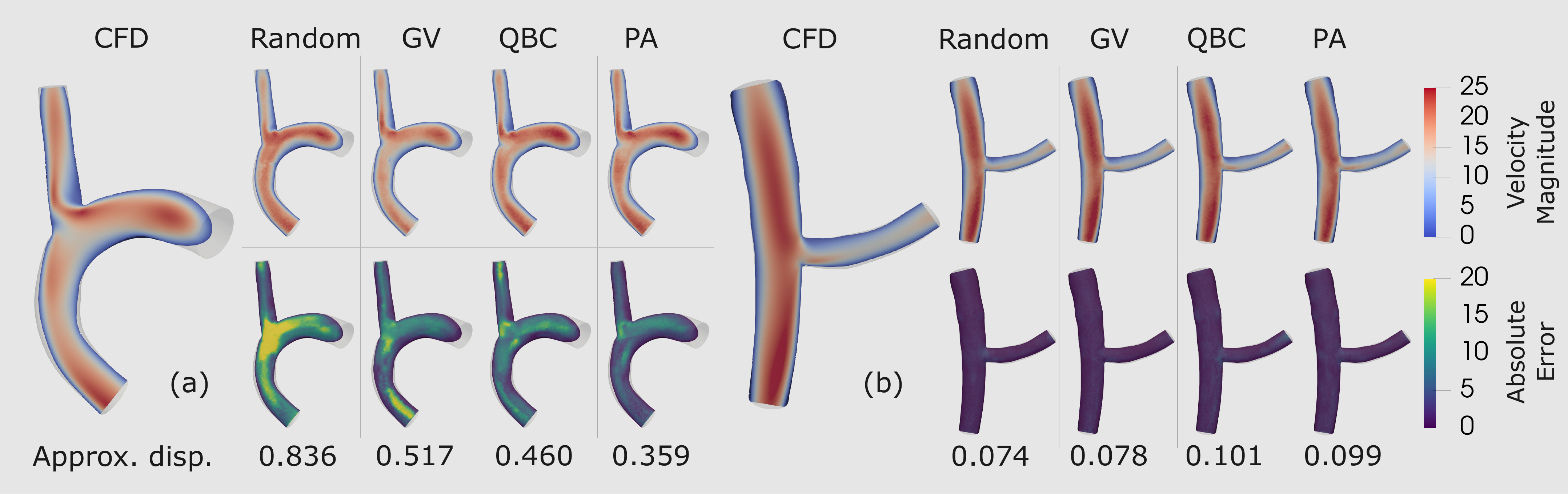}
    \caption{Qualitative comparison of predicted velocity fields on a hard (a) and easy (b) sample for the last round of active learning. The first row shows the magnitude of the predicted velocity field, while the second row the magnitude of absolute error between CFD and respective query strategy.}
    \label{fig:visual}
\end{figure}